\documentclass[sigconf]{acmart}

\AtBeginDocument{%
  \providecommand\BibTeX{{%
    \normalfont B\kern-0.5em{\scshape i\kern-0.25em b}\kern-0.8em\TeX}}}

\copyrightyear{2019}
\acmYear{2019}
\acmConference[KDD '19]{KDD 2019: Workshop on Intelligent Information Feed}{August 04--08, 2019}{Anchorage, Alaska USA}

\usepackage{color}
\usepackage{subfigure}
\usepackage{graphicx}

\begin{document}

\title{News Cover Assessment via Multi-task Learning}

\author{Zixun Sun}
\authornote{Both authors contributed equally to this research.}
\author{Shuang Zhao}
\authornotemark[1]
\affiliation{%
  \institution{Interactive Entertainment Group, Tencent Inc.}
  \streetaddress{No. 1801 Hongmei Rd.}
  \country{China}
  \postcode{200235}
}
\email{zixunsun@tencent.com}
\email{nicholezhao@tencent.com}

\author{Chengwei Zhu}
\affiliation{%
  \institution{Interactive Entertainment Group, Tencent Inc.}
  \streetaddress{No. 1801 Hongmei Rd.}
  \country{China}}
\email{chavezzhu@tencent.com}

\author{Xiao Chen}
\affiliation{%
  \institution{Interactive Entertainment Group, Tencent Inc.}
  \country{China}
}
\email{evelynxchen@tencent.com}

\renewcommand{\shortauthors}{Trovato and Tobin, et al.}

%
\begin{abstract}

Online personalized news product needs a suitable cover for the article. The news cover demands to be with high image quality, and draw readers' attention at same time, which is extraordinary challenging due to the subjectivity of the task.  
In this paper, we assess the news cover from image clarity and object salience perspective. 
We propose an end-to-end multi-task learning network for image clarity assessment and semantic segmentation simultaneously, the results of which can be guided for news cover assessment. 
The proposed network is based on a modified DeepLabv3+ model. The network backbone is used for multiple scale spatial features exaction, followed by two branches for image clarity assessment and semantic segmentation, respectively. 
The experiment results show that the proposed model is able to capture important content in images and performs better than single-task learning baselines on our proposed game content based CIA dataset. 

\end{abstract}

\keywords{Multi-task, Convolutional Neural Network, Image Quality Assessment, Semantic Segmentation}

\maketitle

\section{Introduction}
With the growth of personalized information consumption demand, the recommendation system has achieved significant success in news application (app), such as ``kuaibao" and ``toutiao" in China, which aiming at recommending high quality articles based on individual demand.  
In feed list of those apps, only titles and cover images are exhibited, with which readers decide whether to click and read the articles. 
A high quality cover will significantly increase the article click-through rate (CTR) and improve the readers' quality of experience (QoE) simultaneously. 
Thus the cover image assessment is particularly crucial for information feed design.

In this paper, we focus on the task of cover assessment in game content, such as the apps of Kings' Campsite, which is a professional generated content platform for players of Honor of Kings.
One high quality cover image should be appealing to user's attention and meanwhile express the article content. 
Comparing the quality of images can seem like a very subjective task. What makes one image preferable to another depend on many factors, and may vary depending on the different individuals. 
Intuitively, each news cover is chosen or cropped from the article image content, and demanding to be presented with high clarity and prominent objects. 
Motivated by this, we decompose the challenging task of determining cover images into two subtasks, the image clarity assessment and the image semantic segmentation\cite{mottaghi2014role,cordts2016cityscapes,chen2018deeplab,chen2017rethinking,chen2018encoder}.
The main idea is that the image semantic segmentation results help to quantify the semantic information including object proportion and position in images, which further along with image clarity assessment results perform as news cover selection and cropping guideline.

In recent years, the Convolutional Neural Networks (CNNs) have shown a great progress in computer vision research, especially when the ``AlexNet" appeared and achieved striking results in the ImageNet competition in 2012 \cite{krizhevsky2012imagenet}. 
The CNN performance on classification is further improved in 2014, in which VGGNet \cite{simonyan2014very} and GooleNet \cite{szegedy2015going} with deeper and wider network architecture achieved high performance in the 2014 ILSVRC \cite{russakovsky2015imagenet} classification challenge.
With the advent of ResNet \cite{he2016deep}, the network architectures are going even more deeper and achieved higher performance.

The success of CNNs has bloomed the research of their application to a variety of computer vision tasks, i.e., image quality assessment (IQA) \cite{kang2014convolutional,liu2017rankiqa,bosse2018deep, talebi2018nima} and semantic segmentation\cite{mottaghi2014role,cordts2016cityscapes,chen2014semantic,chen2018deeplab,chen2017rethinking,chen2018encoder}. 
These studies result in significant improvement compared to earlier hand-crafted based features. 
On one hand, semantic segmentation performance applying Deep Convolutional Neural Networks (DCNNs) has been successfully improved, for example the outstanding performance of DeepLab series networks\cite{chen2014semantic,chen2018deeplab,chen2017rethinking,chen2018encoder}, especially DeepLabv3\cite{chen2017rethinking} and DeepLabv3+\cite{chen2018encoder}.
On the other hand, the image quality with respect to human perception can be accurately predicted with CNN-based method \cite{kang2014convolutional,liu2017rankiqa,bosse2018deep, talebi2018nima}.

\begin{figure*}
  \includegraphics[width=0.95\textwidth]{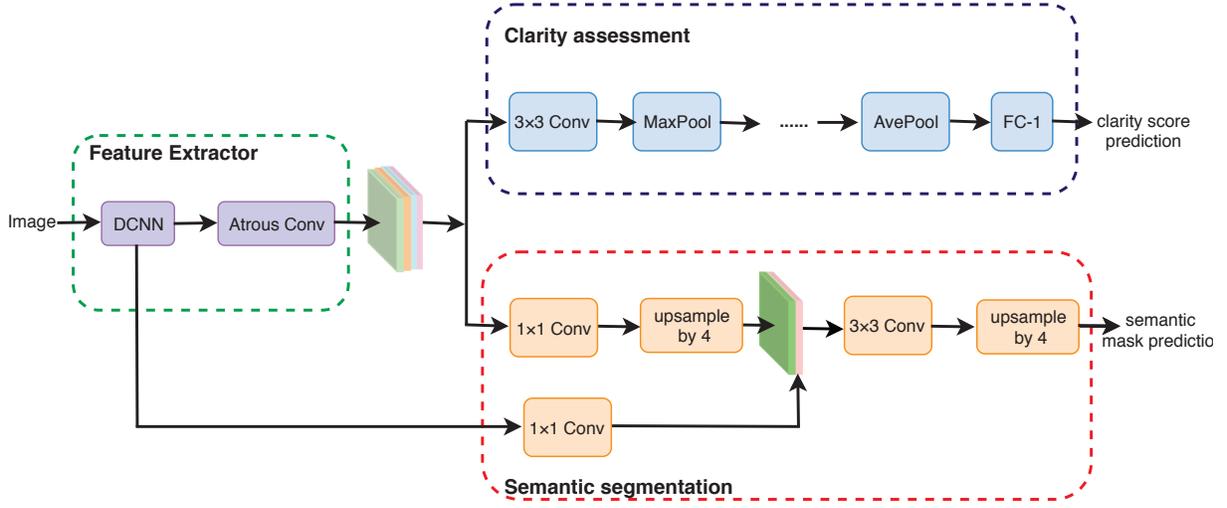}
  \caption{Multi-task learning network architecture.}
  \Description{Our proposed multi-task learning network extends DeepLabv3+ by adding a branch of image clarity assessment module.}
  \label{fig:model}
\end{figure*}

Motivated by the ability of DeepLabv3+ \cite{chen2018encoder} to capture the contextual information at multiple scales, in this paper we employ DeepLabv3+ as the network architecture and model a multi-task learning network to assess news cover from image clarity and semantic segmentation perspective, as illustrated in Fig. \ref{fig:model}. 
We focus on pursuing a practical game content cover image assessment system, which jointly performs image clarity assessment and semantic segmentation prediction. 
Finally, we experimentally verify the effectiveness of the proposed model on our Cover Image Assessment (CIA) dataset. 
Another contribution of our proposed multi-task learning network is that it addresses the limitations of available modern graphics processing units (GPUs) memory when processing multi-task deep learning, in which the memory complexity is independent of the number of tasks in our proposed end-to-end multi-task networks model.

\section{Related Work}
We briefly review the literature related to our approach, and conclude as below.

\textbf{Image quality assessment (IQA)}.
Human visual system is highly sensitive to edge and contour information of an image \cite{martini2012image}. Some IQA studies take edge structure information as the main image quality consideration, for example, in \cite{cohen2010no} the authors apply edge information for both blur and noise detection, which are the major factors on image quality degradation. 
In \cite{ni2016screen}, an edge model is employed to extract salient edge information for screen content images assessment, which outperforms the other state-of-the-art IQA models of the day.

In recent years, the idea of employing a CNN based approach for no-reference IQA (NR-IQA) tasks is arising, and meanwhile the performance of NR-IQA has been significantly improved under such methods\cite{niu2019deeplab,kang2014convolutional}. For example, in \cite{kang2014convolutional}, a CNN is directly utilized for image quality prediction without a reference image, which integrates the feature learning and regression into one optimization process. 
One common ground behind those models is that these network architectures are shallower and narrower, which are not deep enough for learning high-level features. 
The emergence of deeper CNN, such as ResNet-101 \cite{he2016deep} and Xception \cite{chollet2017xception}, further promotes the representational abilities of those models. 
For example, DeepLabv3+ \cite{chen2018encoder}, employs atrous convolution to extract dense feature maps and capture global multiple scale context, resulting in significant performance improvement over semantic segmentation tasks. In \cite{bosse2018deep}, DeepLab based network is applied to excavate spatial features of hyper spectral images, and achieves outstanding performance. 

Most IQA studies consider low-level features such as color or texture, which is not enough for news cover assessment. In news cover assessment, the high-level object feature is also one of the key factors for future cover image selection or cropping. In this paper, we take both low-level image clarity feature and high-level object feature into consideration for news cover assessment.

\textbf{multi-task learning (MTL)}. MTL is based on a fundamental idea that different tasks could share a common low level representation. 
In many computer vision tasks, MTL has exhibited advantages in performance improvement and memory saving. 
In \cite{kokkinos2017ubernet}, one unified architecture which jointly learn low-, mid-, and high-level vision tasks is introduced. With such a universal network, the tasks of boundary detection, normal estimation, saliency estimation, semantic segmentation, semantic boundary detection, proposal generation, and object detection can be simultaneously addressed. 
In \cite{misra2016cross}, a multi-task learning network with "cross-stitch" units is proposed, which shows dramatically improved performance over one-task based baselines on the NYUv2 dataset \cite{silberman2012indoor}. 
However, prior studies have not explored multi-task learning architecture or approach for IQA and semantic segmentation, which is our target method in this work. 


\section{Method}

Inspired by the superior performance in semantic segmentation tasks and the multiple scale spatial features to capture interesting part of an image, DeepLabv3+ network\cite{chen2018encoder} is chosen as the a basis for the proposed multi-task learning based cover image assessment network.
Our proposed network architecture is illustrated in Fig.\ref{fig:model}, which is comprised by the feature extractor module, the image clarity assessment module, the semantic segmentation module.
The proposed multi-task learning network is modified from DeepLabv3+ network by adding an image clarity assessment branch and meanwhile retaining the encoder-decoder architecture in DeepLabv3+ for semantic segmentation tasks. 
The network first processes the whole image with a deep convolutional neural network (DCNN) and subsequently employs atrous convolution to produce a feature map with multi-scale contextual information. 
Then the feature map is shared between clarity assessment module and semantic segmentation module simultaneously.

\subsection{Deep image clarity assessment learning}

The proposed multi-task learning network has two output layers. The first layer outputs the image clarity assessment score $\hat{y}_k$, for the image indexed by $k$. The second layer outputs a predicted binary mask matrix to distinguish foreground and background in an image. 

In image clarity assessment module, the feature map is regressed by a sequence of conv3-256, maxpool, conv3-256, maxpool, conv3-256, maxpool, one average pooling layer and one FC-1 layer. The convolutional layers apply $3 \times 3$ pixel-size convolution kernels and are activated by a rectified linear unit (ReLU). The max pooling layers apply $2 \times 2$ pixel-size kernels. The output of FC-1 layer is activated by a sigmoid unit. The estimated image clarity score $\hat{y}_k$ of image $k$ is then obtained. 
Note that the image clarity assessment is a regression task, we choose Mean Square Error (MSE) as the loss function for training image clarity assessment branch, which is 
\begin{equation}
	L_{c}^k = (y_k - \hat{y}_k)^2.
\end{equation}

\subsection{Deep semantic segmentation learning}
\label{sec_segm}

For the task of semantic image segmentation, our proposed multi-task learning network keeps the encoder-decoder architecture in DeepLabv3+. 
In this paper, the feature map from feature exactor is computed with $output$ $stride = 16$, which is consistent with DeepLabv3+. 
After $1\times 1$ convolution operation and one simple bilinear upsampling with factor 4, the feature map concatenates with the low level features from the DCNN and then is refined by a few $3 \times 3$ convolutions. Finally, the refined feature map is  upsampled by another simple bilinear upsampling with factor 4 and is then feeded into semantic segmentation prediction.  

Note that the image with low clarity may introduce noise during semantic segmentation network training.
Denote $x_k^{(i,j)}$ as the binary annotation of $(i,j)$-th pixel of image $k$. A ground truth label $x_k^{(i,j)} = 1$ means foreground (object) for $(i,j)$-th pixel and $0$ for background. 
Denote $\hat{x}_k^{(i,j)}$ as the predicted pixel mask value of image $k$.
The cross entropy loss function $L_{s}^k$ is applied for semantic segmentation, 
\begin{equation}
	L_{s}^k = -\sum_{i,j} \big[x_{k}^{(i,j)} \cdot \text{log}\hat{x}_{k}^{(i,j)} + (1-x_k^{(i,j)})\text{log}(1-\hat{x}_k^{(i,j)}) \big].
\end{equation}



\subsection{End-to-End multi-task learning network}
With above definitions, we minimize an objective function following the multi-task loss in our proposed network model. 
To avoid the influence of heavy distorted images, we train the semantic segmentation branch over the images with relatively high clarity, i.e., image clarity score $y_k > 2.3$ in this paper.
Our multi-task loss function is defined as:
\begin{equation}\label{multi-task-loss}
	L = \frac{1}{N}\sum_{k=1}^{N}\big[L_{c}^k + \lambda \alpha_k L_{s}^k\big],
\end{equation}
where $N$ denotes the number of training images, the hyper-parameter $\lambda$ control the balance between two task losses, and $\alpha_k$ is the binary indicator with 
$$ \alpha_k = \left\{
\begin{aligned}
	&1& \ \text{if} \  y_{k} > 2.3 \\
	&0& \ \text{otherwise.} 	
\end{aligned}
\right.
$$
By convention the image with low clarity is labeled $\alpha_k = 0$ and hence the corresponding $L_{s}^k$ is ignored.

\section{Experimental Results}

In this section, we introduce our dataset and conduct a number of experiments to evaluate the performance of our approach.

\subsection{Cover Image Assessment Dataset}

\begin{figure*}[ht]
	\centering
    \subfigure[Image clarity score = 9.5]{
		\centering
		\includegraphics[width=0.45\textwidth]{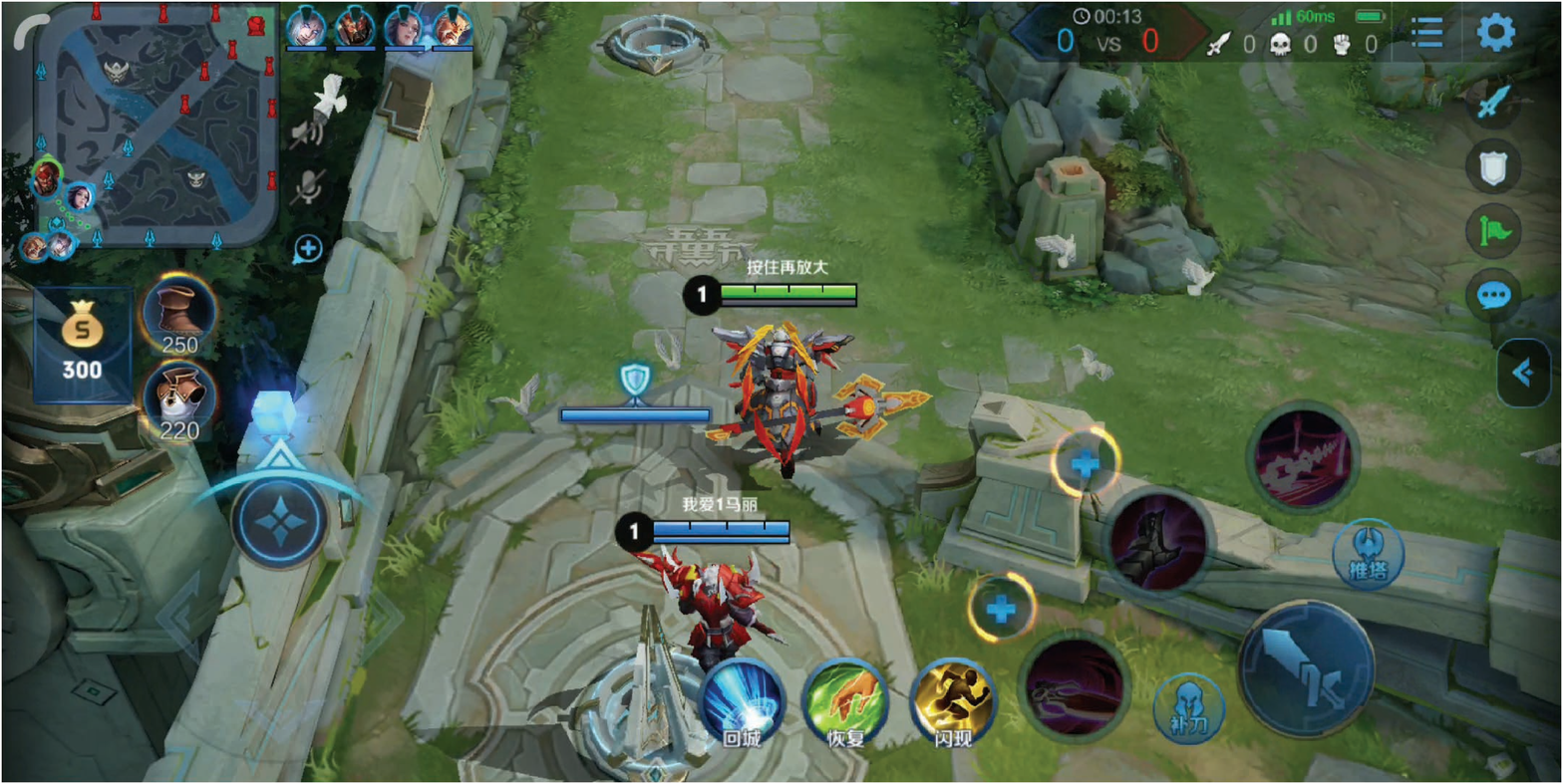}
		\label{fig:img_9.5}
	}
	\subfigure[Image clarity score = 7.5]{
		\centering
		\includegraphics[width=0.45\textwidth]{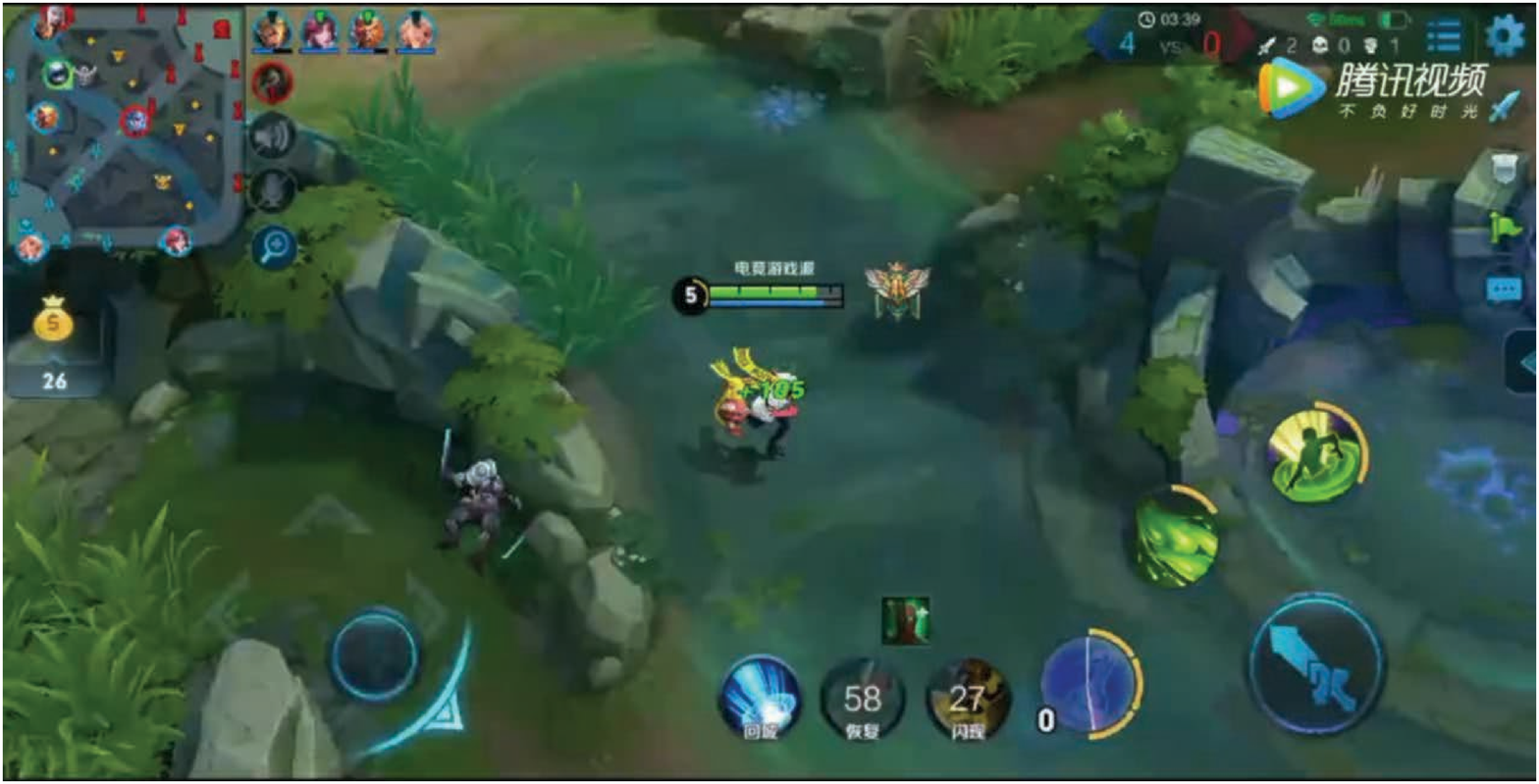}
		\label{fig:img_7.5}
    }
    \subfigure[Image clarity score = 4.22]{
        \centering
        \includegraphics[width=0.45\textwidth]{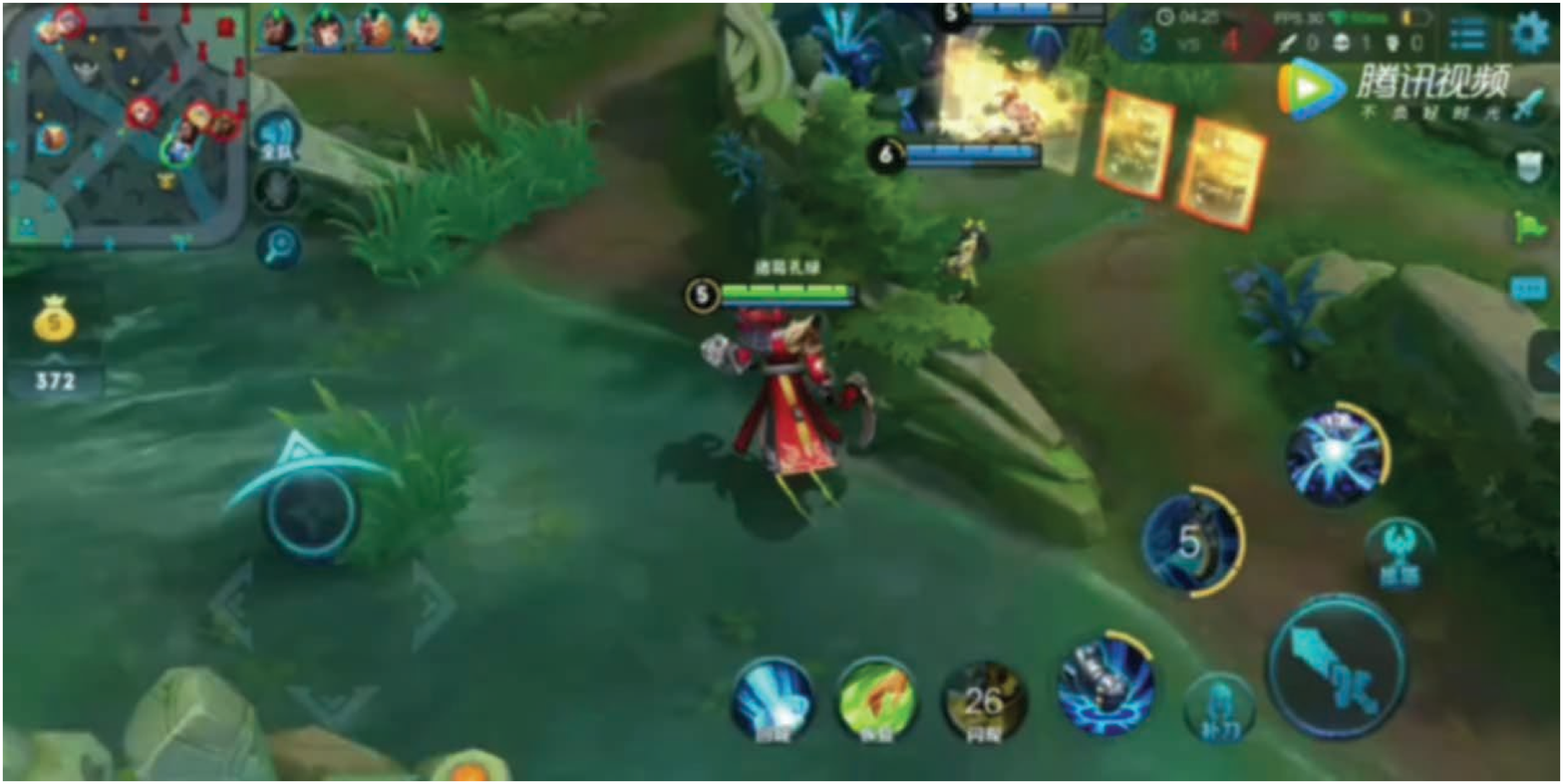}
        \label{fig:img_4.22}
    }
    \subfigure[Image clarity score = 1.88]{
        \centering
        \includegraphics[width=0.45\textwidth]{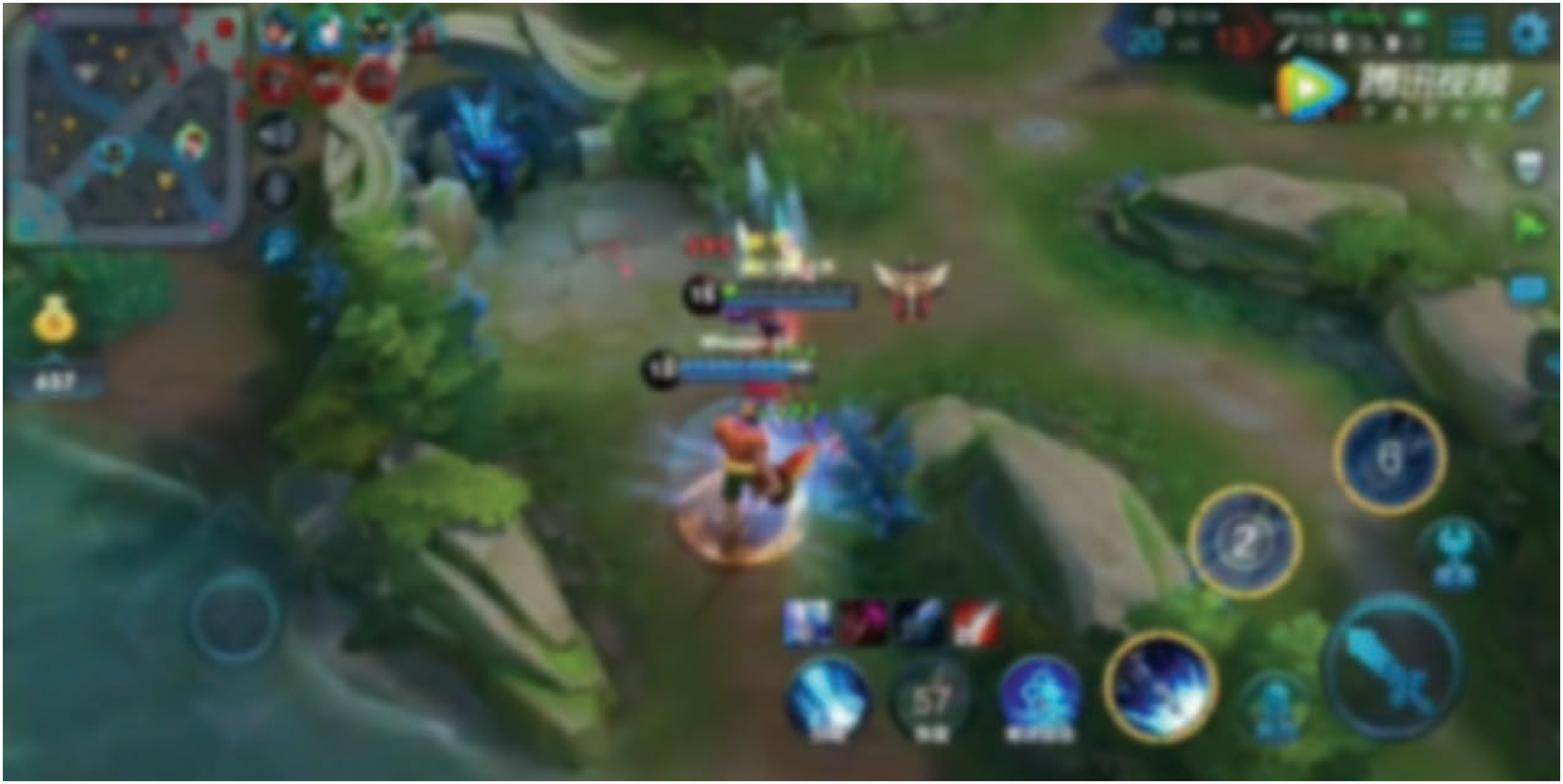}
        \label{fig:img_1.88}
    }    
	\centering
    \caption{Some example images from CIA dataset with image clarity score}
    \label{fig:distorted image example}
\end{figure*}

We experiment on Cover Image Assessment Dataset (CIA), which is consisted by game images from Honor of Kings, a multiplayer mobile online battle game. 
Each image in the dataset is annotated from pixel-level and meanwhile labeled by one clarity score. 
Distinguished from the previous image quality assessment datasets, all of which provide the visual quality of distorted images, CIA dataset assesses the image from two dimension, image clarity and semantic information, which is more practical for cover image assessment. 
Next, we introduce the detailed collection process of this dataset and provide its statistics results. 

We collect a batch of original Honor of Kings game videos from Tencent video website, the resolution of which consisted by  $1280 \times 720$ and $2248 \times 720$. 
All the videos are of high quality and the clarity among them are consistent. 
We then extract 1021 frames from those videos, and the extracted images contain 98 objective hero classes and one background class. 
Our annotation tasks include two aspects. One is pixel-level mask annotation for semantic segmentation task. We annotate the objective heros in images.
Another is image clarity assessment. We distort each image to 10 levels through Gaussian blur distortion and quantize corresponding clarity score to range $[1,10]$. Higher value of clarity score (score 10 is maximal) corresponds to higher image clarity. 
A few examples with ratings associated with different levels of clarity are illustrated in Fig. \ref{fig:distorted image example}.  
Thus through distortion process our CIA dataset is augmented to 8651 images, and each image has one clarity score and one segmentation mask.

\subsection{Experimental Setup}

We investigate a number of network backbones and training strategies, and evaluate their performance through standard semantic segmentation metrics and clarity assessment metrics. 
The network backbone is based on popular deep models such as MobileNet \cite{sandler2018mobilenetv2}, Xception \cite{chollet2017xception}, ResNet-101\cite{he2016deep}, and is trained on our proposed CIA dataset. 
The dataset is randomly split into 7786 images for training set and 865 images for testing set. In both training and testing dataset, the proportion of distorted image and undistorted image is $9:1$, which guarantees that the undistorted image has been seen by the network during training. Random crop is employed during the training as a data enhancement method, and the cropped image size is $513 \times 513$. We use the Pytorch framework and employing Stochastic Gradient Descent (SGD) with an initial learning rate of $7 \times 10^{-3}$ and momentum of $0.9$ for network training. 
The models are trained for $50$ epochs, which ensures the convergence of all models. 
The batch size is set to $8$ and the parameter $\alpha_k$ is set to $2.3$ during both training and testing. The parameter $\lambda$ is set to $0.1$. Training rates follow a polynomial decay policy with factor $2$. 

In this paper, we train the multi-task learning based network in two strategies, introduced as end-to-end training and multi-stage training. 
Specifically, in end-to-end training, the tasks of  image clarity assessment and semantic segmentation are trained simultaneously, with the multi-task loss \eqref{multi-task-loss} be minimized in each iteration. 
In multi-stage training, the semantic segmentation branch is trained first and the convergent feature map from feature extractor is shared for clarity assessment model. We then fine-tune the following convolutional layers and fully-connected layers with mini-batches of 8.

We compute the average Pearson Linear Correlation Coefficient (LCC) and mean Intersection over Union (mIoU) as evaluation metrics for task of image clarity assessment and task of semantic segmentation, respectively. 
The LCC is computed as below: 
\begin{equation}
	LCC = \frac{\sum_1^N(y_k-\overline{y})(\hat{y}_k - \overline{\hat{y}})}{\sqrt{\sum_1^N(y_k - \overline{y})^2}\sqrt{\sum_{i=1}^N(\hat{y}_k - \overline{\hat{y}})^2}},
\end{equation}
where $\overline{y}$ and $\overline{\hat{y}}$ are the means of ground truth and predicted clarity score, separately. LCC measures the linear correlation between the predicted image clarity score and the ground truth, and larger LCC value implies more accurate image clarity assessment performance. 
The mIoU is defined over image pixels, following the standard protocols \cite{everingham2010pascal}. 
Large mIoU means higher accuracy on semantic segmentation performance. With semantic segmentation results, we can compute corresponding semantic information easily, which along with predicted image clarity score help for news cover assessment.


\subsection{Performance Evaluation}

We first evaluate our proposed multi-task learning network performance over several network backbones on CIA dataset, and then further demonstrate the effectiveness of our proposed multi-task learning model through comparing with single-task models. 

\begin{figure*}[ht]
	\centering
    \subfigure[Image clarity score: 8.2036(9.5), IoU = 0.8369, Object proportion: 1.49\%]{
		\centering
		\includegraphics[width=0.45\textwidth]{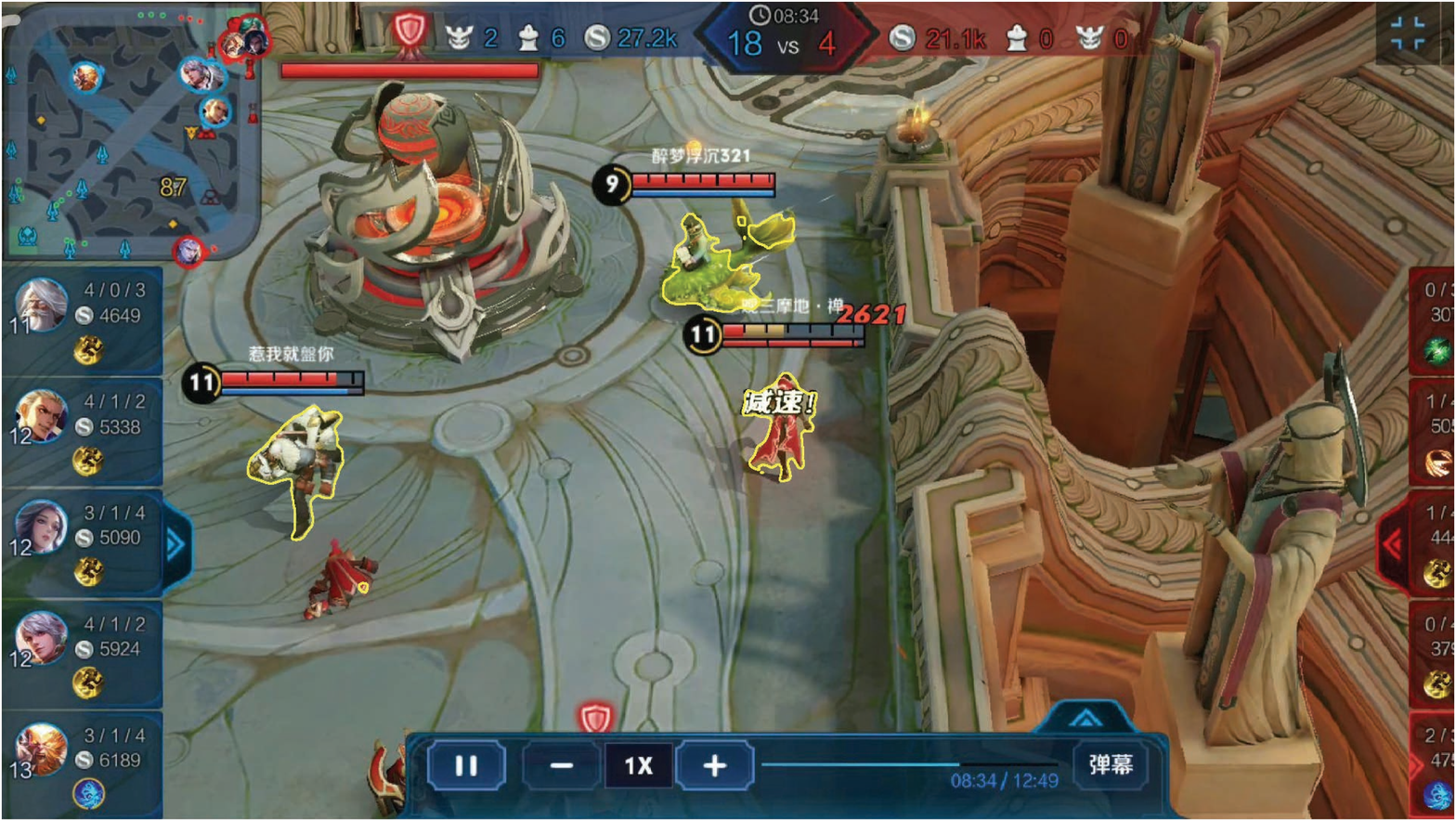}
		\label{fig:img_a}
	}
	\subfigure[Image clarity score = 5.7954(5.625), IoU = 0.7534, Object proportion: 1.57\%]{
		\centering
		\includegraphics[width=0.45\textwidth]{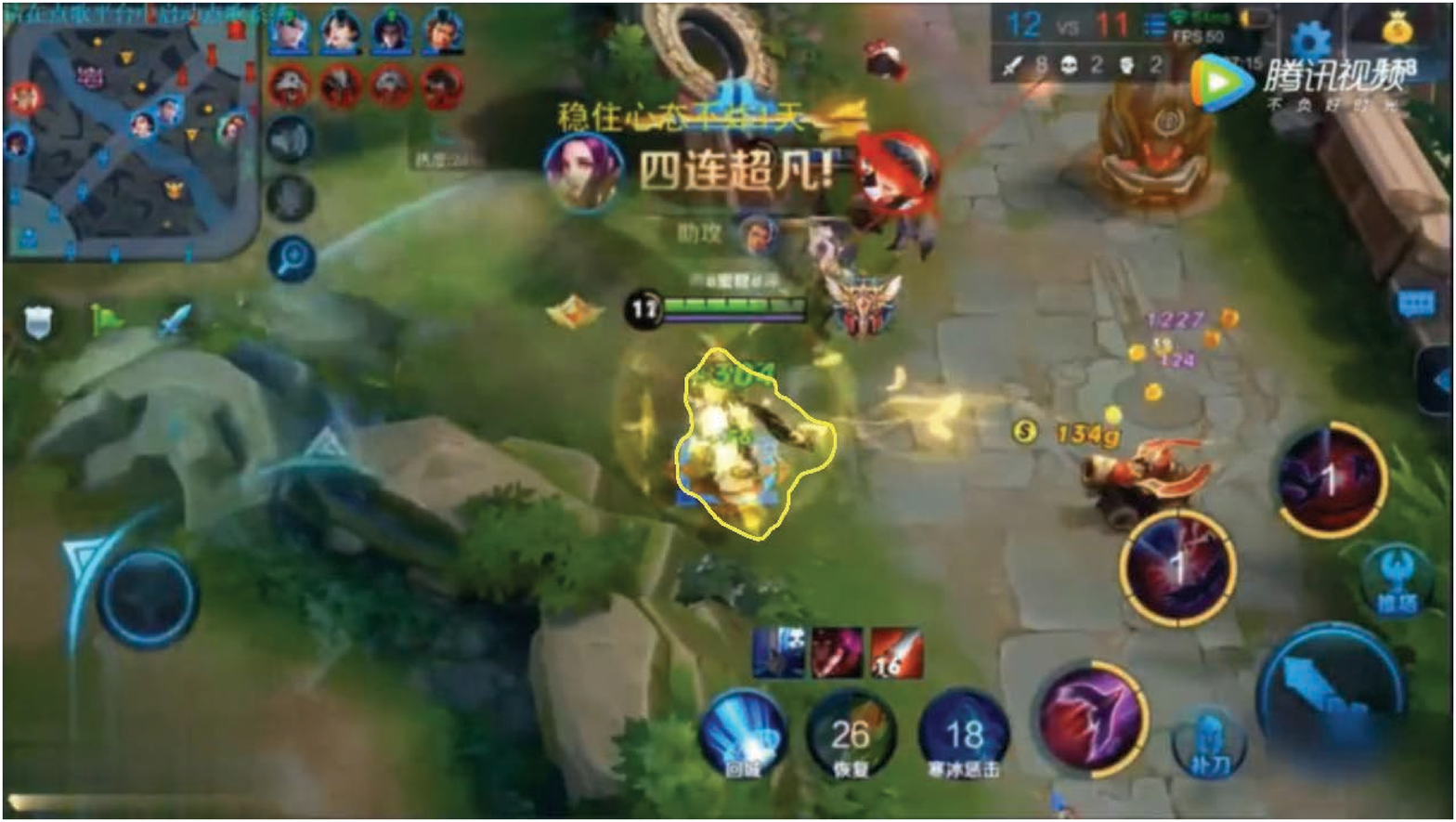}
		\label{fig:img_b}
    }
    \subfigure[Image clarity score = 4.5992(4.2188), IoU = 0.7514, Object proportion: 2.13\%]{
        \centering
        \includegraphics[width=0.45\textwidth]{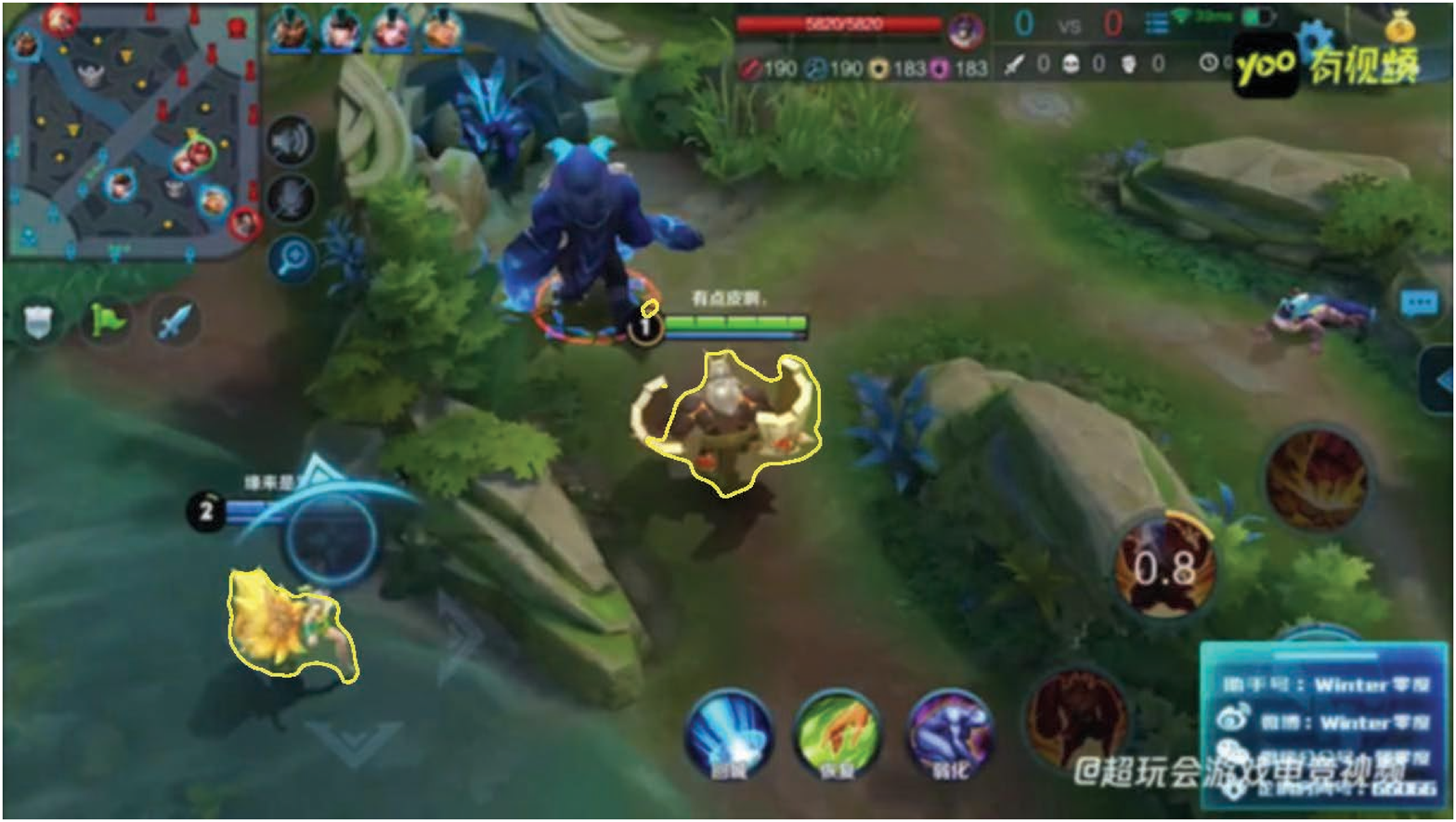}
        \label{fig:img_c}
    }
    \subfigure[Image clarity score = 2.3295(2.31), IoU = 0.7518, Object proportion: 2.42\%]{
        \centering
        \includegraphics[width=0.45\textwidth]{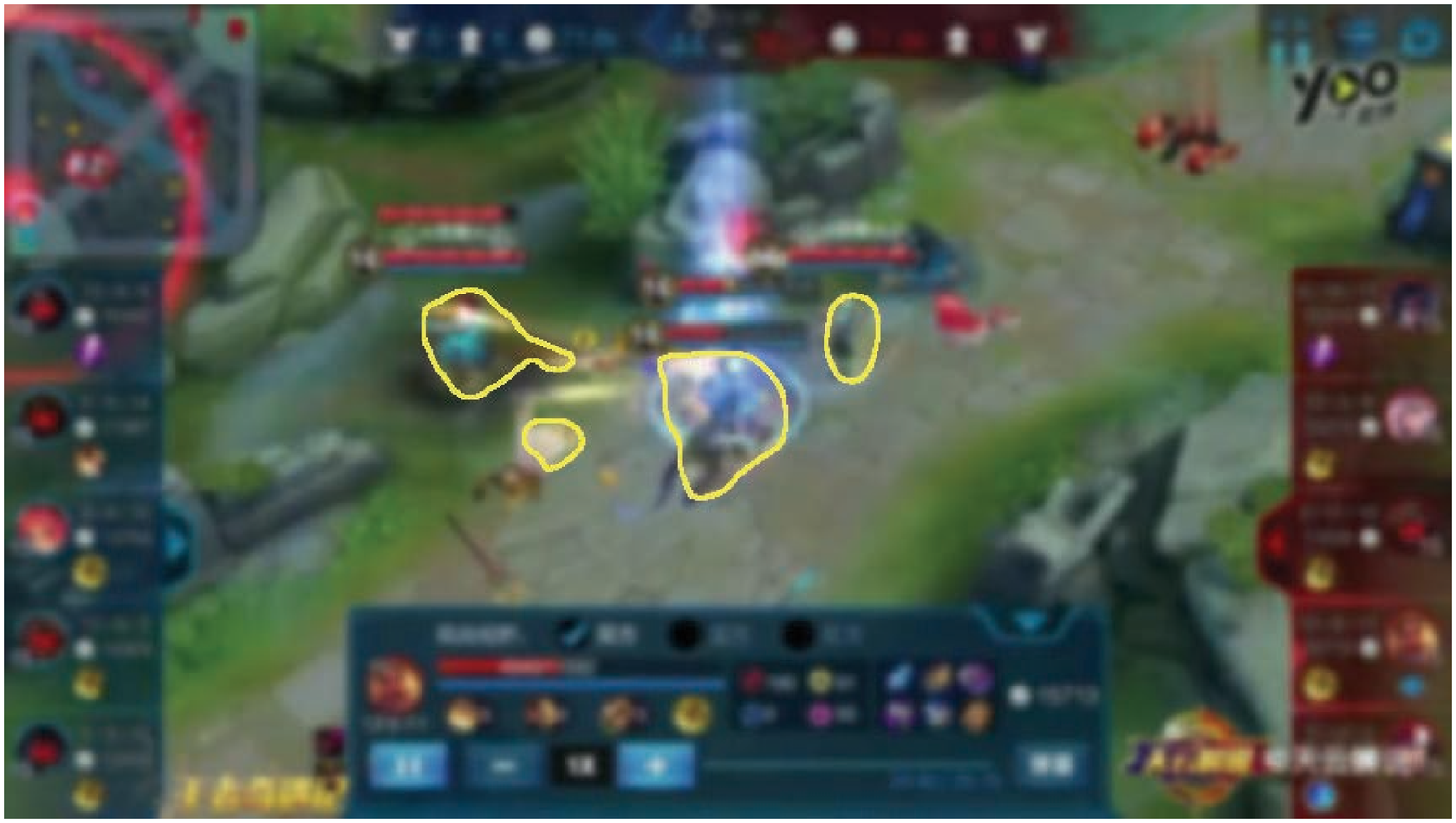}
        \label{fig:img_d}
    }    
	\centering
    \caption{Some example images from CIA dataset using our multi-task network structure. Predicted image clarity (and ground truth) scores, predicted IoU and corresponding object proportion results are shown below each image.}
    \label{fig:experiment results}
\end{figure*}

\textbf{Effectiveness of Network Structure.}
To demonstrate the ability of our proposed multi-task learning network structure to assess the image clarity and predict semantic segmentation simultaneously, we train the multi-task learning network on CIA data set and compare different network backbone performance among MobileNet, Xception, and ResNet-101. The network is trained in end-to-end manner. The network backbones are initialized with a pre-trained weights training on ImageNet. 
Fig. \ref{fig:experiment results} shows a few predicted results from CIA dataset under our proposed multi-task learning network. 
The segmentation results are visualized in Fig. \ref{fig:img_a}-\ref{fig:img_d}, and the predicted image clarity scores, ground truth scores, predicted IoU and corresponding object proportion in images are shown below each image. 
The object proportion is computed by corresponding semantic segmentation. 
Results in Fig. \ref{fig:experiment results} suggest that our proposed multi-task learning network works, and the network has the ability to predict the image clarity and semantic segmentation simultaneously, which helps to design cover image selection and cropping strategies. 
The detailed performance results under different network backbones are summarized in Table \ref{tab:diff-backbone}.
As can be seen, ResNet-101 performs better than the other two network backbones, in terms of LCC and mIoU. 
Based on these results, we choose ResNet-101 as our network structure in the following studies.

\begin{table}
    \caption{Performance evaluation on CIA dataset in terms of different network backbones}
    \label{tab:diff-backbone}
    \begin{tabular}{lccc}
        \toprule
        Type/Network & MobileNet & Xcpetion & ResNet\\
        \midrule
        LCC & 0.9809 & 0.9559 & 0.9805 \\
        mIoU & 0.7603 & 0.7165 & 0.7746 \\
        \bottomrule
    \end{tabular}
\end{table}

\begin{table}
    \caption{Performance comparison among different models}
    \label{tab:diff-model}
    \begin{tabular}{lccc}
        \toprule
        Models & Image clarity & Semantic segmentation & Multi-task\\
        \midrule
        LCC & 0.9741 & $/$ & 0.9805\\
        mIoU & $/$ & 0.7618 & 0.7746\\
        \bottomrule
    \end{tabular}
\end{table}

\textbf{Effectiveness of multi-task learning.}
Next, to further demonstrate whether the proposed multi-task learning model has the potential to improve results, we train the baseline network with image clarity assessment branch only, and with semantic segmentation branch only. 
The comparison results are shown in Table \ref{tab:diff-model}. 
The column \emph{image clarity} and column \emph{semantic segmentation} in  Table \ref{tab:diff-model} are single-task learning model. 
Obviously, across all three types of network, our proposed multi-task learning network outperforms both single-task based semantic segmentation model and clarity assessment model, in terms of mIoU and LCC respectively.
This verifies the effectiveness of our proposed multi-task learning models.
Besides, for multi tasks learned with several single-task networks, the demand for memory for back-propagation algorithm intensifies linearly with the number of tasks. Instead, under our proposed end-to-end multi-task networks model, the memory complexity is independent of the number of tasks. This indicates that the proposed multi-task learning network has advantages in fulfilling multi-task and multi-target with a limited memory budget.

\textbf{Comparison between two different training strategies}
Finally, we evaluate the impact of different training strategies on the proposed multi-task learning based model. 
Instead of training the multi-task learning network end-to-end, we also evaluate the performance of the proposed network through multi-stage training. In multi-stage training, the clarity assessment module is fine-tuned based on the multiple scale spatial feature map from pre-trained image segmentation network on CIA dataset.
The comparison results are presented in table \ref{tab:diff-training}. Obviously, the end-to-end training method leads to both excellent image clarity assessment and segmentation performance.


\begin{table}
    \caption{Performance comparison among different training strategies}
    \label{tab:diff-training}
    \begin{tabular}{lccc}
        \toprule
        Training strategies & end-to-end training & multi-stage training\\
        \midrule
        LCC & 0.9805 & 0.8548\\
        mIoU & 0.7746 & 0.7470\\
        \bottomrule
    \end{tabular}
\end{table}

\section{Conclusion}

News cover assessment is the critical step for suitable cover photo generation. 
To address challenges in news cover assessment, in this work, we have introduced an end-to-end multi-task learning network, which can predict image clarity and semantic segmentation in parallel, and reduce the memory consumption compared to task execution independently.
Then various news cover selection and cropping strategies can be designed with image clarity assessment and semantic information on salient object. We have experimentally shown the effectiveness of the proposed multi-task learning network on our proposed game content based CIA dataset. 
How to design news cover selection and cropping strategy according to the predicted image clarity and semantic information on salient object is the future main research direction. 
Research in these directions is underway, and more importantly, we consider the work in this paper to be a first step in the way of news cover assessment.

\bibliographystyle{ACM-Reference-Format}
\bibliography{multi-task}

\end{document}